\documentclass[10pt,twocolumn,letterpaper]{article}

\usepackage[pagenumbers]{iccv} 

\definecolor{iccvblue}{rgb}{0.21,0.49,0.74}
\usepackage[pagebackref,breaklinks,colorlinks,allcolors=iccvblue]{hyperref}
\usepackage{multirow}

\def\paperID{10886} 
\def\confName{ICCV}
\def\confYear{2025}

\title{PointGS: Point Attention-Aware Sparse View Synthesis with Gaussian Splatting}

\author{Lintao Xiang, Hongpei Zheng, Yating Huang, Qijun Yang, Hujun Yin\\
Department of Electrical and Electronic Engineering,\\
The University of Manchester
}

\begin{document}
\maketitle
\begin{abstract}
3D Gaussian splatting (3DGS) is an innovative rendering technique that surpasses the neural radiance field (NeRF) in both rendering speed and visual quality by leveraging an explicit 3D scene representation. Existing 3DGS approaches require a large number of calibrated views to generate a consistent and complete scene representation. When input views are limited, 3DGS tends to overfit the training views, leading to noticeable degradation in rendering quality. To address this limitation, we propose a Point-wise Feature-Aware Gaussian Splatting framework that enables real-time, high-quality rendering from sparse training views. Specifically, we first employ the latest stereo foundation model to estimate accurate camera poses and reconstruct a dense point cloud for Gaussian initialization. We then encode the colour attributes of each 3D Gaussian by sampling and aggregating multiscale 2D appearance features from sparse inputs. To enhance point-wise appearance representation, we design a point interaction network based on a self-attention mechanism, allowing each Gaussian point to interact with its nearest neighbors. These enriched features are subsequently decoded into Gaussian parameters through two lightweight multilayer perceptrons (MLPs) for final rendering. Extensive experiments on diverse benchmarks demonstrate that our method significantly outperforms NeRF-based approaches and achieves competitive performance under few-shot settings compared to the state-of-the-art 3DGS methods.
\end{abstract}    
\section{Introduction}
\label{sec:intro}

Novel view synthesis (NVS) has been a longstanding and intricate challenge in the field of computer vision, with the objective of rendering unseen viewpoints from a collection of images with known camera parameters. This task is fundamental to a range of applications such as gaming, VR, AR, cultural heritage preservation, robotics and autonomous driving. Neural Radiance Field (NeRF)~\cite{mildenhall2021nerf} has made substantial advances to NVS, enabling high-fidelity renderings by representing scenes as continuous implicit functions of positions and orientations. Although NeRF-based methods~\cite{barron2021mip, barron2022mip, muller2022instant} have demonstrated great potentials, they often result in significant computational demands, negatively impacting on real-time performance. Recently 3D Gaussian Splatting (3DGS)~\cite{kerbl20233d} employs anisotropic 3D Gaussians for explicit scene representation, facilitating efficient rendering via a parallel splatting pipeline. This method shows superior performance compared to the prior NeRF-based methods, in terms of rendering speed and quality. However, both NeRF and 3DGS require a substantial number of densely sampled, well-distributed and calibrated images to achieve effective optimization, which poses limitations to its practical applicability in scenarios where acquiring such data is challenging. 

Therefore, there has been growing interest in few-shot novel view synthesis within the community. In few-shot NeRF, geometric cues such as depth and normal priors are frequently leveraged as additional constraints to optimize implicit neural radiance field. In~\cite{wei2021nerfingmvs} and~\cite{roessle2022dense}, dense depth priors are leveraged to constrain the NeRF optimization. Such prior information is generated by certain pre-trained models such as monocular depth network~\cite{bhat2023zoedepth} or the SfM pipeline, e.g. COLMAP~\cite{schonberger2016structure}. Besides, in ~\cite{jain2021putting} and~\cite{yang2023freenerf} semantic consistency and frequency annealings are utilized to improve the rendering quality of few-shot NeRF. Recently, few-shot settings have also been explored in the context of 3DGS. But the sparse input views typically cover only a limited portion of the scene, and the resulting point clouds generated by COLMAP often exhibit incomplete or fragmented structures. When used to initialize 3DGS, these imperfect reconstructions can lead to suboptimal optimization and floating artifacts in novel view rendering. Although several methods~\cite{chung2024depth, zhu2023fsgs, xiong2023sparsegs} attempt to mitigate this challenge by incorporating various geometric priors—such as surface normals, depth, and semantic information—the improvements in rendering quality remain limited.

In this work, we present a novel framework that incorporates learnable point-wise features to significantly boost the rendering quality of 3DGS under few-shot settings. Specifically, we first utilize the latest stereo foundation model, VGGT~\cite{wang2025vggt}, to estimate accurate camera poses and generate dense 3D point clouds from sparse input views. Meanwhile, multi-scale image features are extracted from each input view using a Feature Pyramid Network (FPN). Each 3D point is then projected onto the corresponding image planes to sample multi-view features. By aggregating these 2D features across views, we associate each 3D point with a learnable point-wise feature representation. To facilitate better scene understanding, we design a point-wise feature interaction network that enables each point to exchange appearance information with its neighboring points, thereby generating enriched feature representations. These encoded point-wise features are then decoded into 3D Gaussian attributes for scene rendering using two lightweight multilayer perceptrons (MLPs). In addition, we introduce a depth smoothness loss to improve the accuracy and continuity of the rendered depth maps. Extensive experiments across multiple datasets demonstrate that our method effectively suppresses floating artifacts and enables more accurate and detailed scene reconstruction. In summary, our main contributions are as follows:

(1) A point-wise feature-aware 3DGS framework is proposed to learn rich appearance features for each Gaussian point by aggregating multi-view image information.

(2) A point-wise feature interaction network based on self-attention is designed, enabling each point to exchange information with its neighbors and generate enhanced feature representations. Besides, a depth smoothness loss is used to regularize scene geometry and improve rendering quality.

(3) Extensive experiments on multiple datasets are performed, demonstrating that our method significantly outperforms NeRF-based approaches and achieves performance comparable to state-of-the-art 3DGS methods under sparse-view settings.
\section{Related Work}
\label{sec:work }

\subsection{Neural Radiance Fields}

Novel view synthesis (NVS) aims to reconstruct unseen viewpoints of a scene or object from a set of input images. Neural Radiance Fields (NeRF)~\cite{mildenhall2021nerf} represent a major breakthrough by modeling scenes as continuous implicit functions of spatial positions and viewing directions, combined with differentiable volume rendering for high-quality view synthesis.

Building on NeRF, follow-up studies have improved rendering performance~\cite{barron2021mip, barron2022mip360}, enhanced generalization~\cite{chen2021mvsnerf}, and extended NeRF to tasks such as pose estimation~\cite{zhang2023pose} and dynamic scene modeling~\cite{huang2022hdr}. However, NeRF typically suffers from long training and inference times.

Although several methods aim to accelerate NeRF~\cite{muller2022instant, fridovich2022plenoxels}, they often sacrifice rendering quality, especially at high resolutions. Recently, 3D Gaussian Splatting (3DGS)~\cite{kerbl20233d} has emerged as a promising alternative, offering real-time rendering and faster training by explicitly modeling scenes with anisotropic 3D Gaussians and using parallel differentiable splatting. Each Gaussian in 3DGS encodes learnable geometry and appearance attributes, which are projected onto image planes via splatting and blended to produce rendered images. Building on this foundation, some works have explored model compression for efficient 3D representation~\cite{fan2023lightgaussian, lee2024compact, liu2024compgs}, large-scale scene reconstruction via divide-and-conquer~\cite{lin2024vastgaussian, liu2024citygaussian}, and open-vocabulary understanding with semantic Gaussians~\cite{guo2024semantic}.

\subsection{Few Shot Novel View Synthesis}

The original NeRF requires lots of calibrated views to optimize its implicit scene representation, which increases computational cost and limits practical deployment. To address this, previous works have explored few-shot NeRF approaches, typically categorized into two types: those that introduce additional constraints (e.g., semantic consistency~\cite{jain2021putting}, geometry regularization~\cite{niemeyer2022regnerf}), and those that incorporate depth priors~\cite{deng2022depth, wang2023sparsenerf} to guide training with limited views. With the emergence of 3DGS as a faster and higher-quality alternative to NeRF, several studies have extended it to few-shot settings. These methods employ depth priors~\cite{li2024dngaussian, paliwal2024coherentgs}, Gaussian unpooling~\cite{zhu2023fsgs}, or floater suppression~\cite{xiong2023sparsegs} to improve scene reconstruction. However, monocular depth priors often offer only coarse guidance and struggle with complex geometry. Other approaches~\cite{chen2021mvsnerf,xu2024murf,charatan2024pixelsplat,chen2024mvsplat} rely on pre-trained generalizable models with per-scene optimization, which is often time-consuming.

\begin{figure*}[htbp]
\begin{center}
    \includegraphics[width=1.0\textwidth]{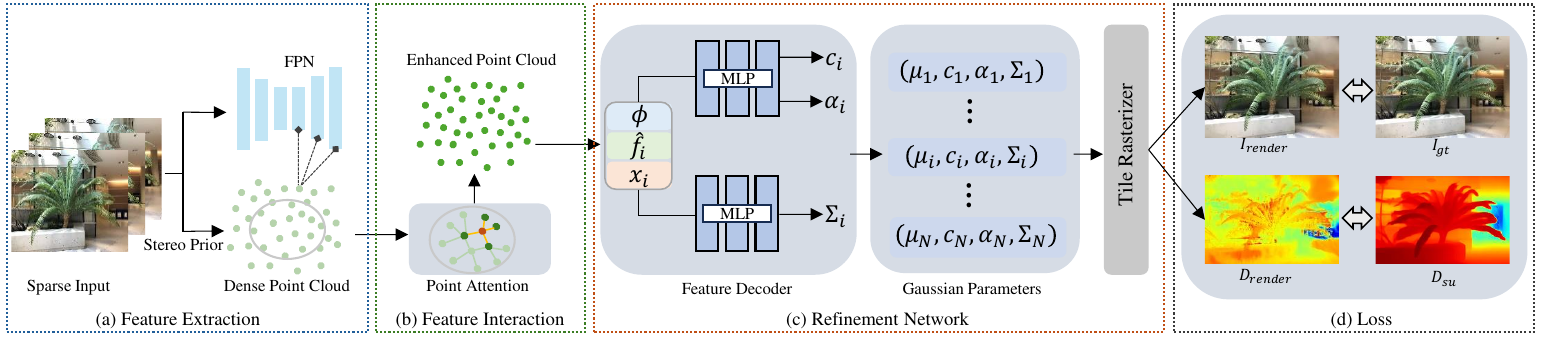}\\
    \caption{\textbf{Overview of the proposed method.} Given a few input images, we firstly generate dense point clouds from VGGT~\cite{wang2025vggt} and extract multi-scale feature maps via an FPN. Each point is then assigned point-wise appearance features by sampling from these feature maps. To enhance feature representation, a self-attention-based interaction module enables information exchange among neighboring points. Finally, two lightweight MLPs decode the 3D Gaussian attributes using the 3D coordinates, view direction, and enhanced features. The overall loss includes photometric and depth regularization terms, along with depth smoothness.}     
    \label{fig:pipeline}
\end{center}
\end{figure*}
\subsection{Preliminary}

3D Gaussian Splatting~\cite{kerbl20233d} is a cutting-edge technique for novel view synthesis, achieving high-quality and real-time free-view rendering by a differentiable rasterization. 3DGS represents scene explicitly with a set of anisotropic 3D Gaussians, each of which is parametrized by its center position $\mu \in \mathbb{R}^3 $ and covariance matrix $\Sigma \in \mathbb{R}^{3\times3}$. The basis function of each Gaussian primitive is defined as:

\begin{equation}
G(x) = e^{-\frac{1}{2}(x-\mu)^T \Sigma^{-1}(x-\mu)} 
\end{equation}

To ensure the positive semi-definiteness of the covariance matrix $\Sigma$ during optimization, $\Sigma$ is further decomposed into a learnable scaling matrix $S$ and rotation matrix $R$ by $\Sigma = RSS^TR^T$. $R$ and $S$ are stored as 3D vector $s \in \mathbb{R}^{3}$ and quaternion $r \in \mathbb{R}^{4}$, respectively. Each Gaussian also includes a learnable opacity $o$ and appearance feature represented by $n$ spherical harmonic (SH) coefficients $\{c_i \in \mathbb{R}^3|i=1,...,n\}$, where $n=D^2$ is the number of coefficients of SH with degree $D$. 

To render an image from a given viewpoint, these 3D Gaussians are firstly splatted from 3D space into a 2D image plane through a viewing transformation $W$ and a Jacobian matrix $J$ representing the affine approximation of the projective transformation. The covariance matrix $\Sigma^{'} \in \mathbb{R}^{2\times2}$ in 2D space can be expressed as:

\begin{equation}
{\Sigma^{'}  = JW \Sigma W^{T}J^{T}}
\end{equation}

Subsequently, the color of each pixel $\mathbf{p}$ is determined by blending $N$ ordered Gaussians $\{G_i | i = 1, \dots, N\}$ that overlap with pixel $\mathbf{p}$ as follows:
\begin{equation}
    C=\sum_{i=0}^{N}c_i\alpha _i\prod_{j=1}^{i-1}(1-\alpha _j)  
\end{equation}
where $c_i$ is the learned color of each Gaussian point, computed by spherical harmonics (SH) coefficients. The density $\alpha_i$ is obtained from the multiplication of 2D covariance matrix $\Sigma^{'}$ and a learned opacity $o_i$ from $G_i$. Gaussians overlapping with pixel $\mathbf{p}$ are sorted in ascending order based on their depths from the current viewpoint. Besides, depth map can be obtained by using the same rasterization pipeline to render the depth values of Gaussian splats.

\begin{equation}
    D=\sum_{i=0}^{N}d_i\alpha _i\prod_{j=1}^{i-1}(1-\alpha _j)  
\end{equation}
where $D$ is the rendered depth and $d_i$ is the the depth of each splat under the camera coordinate system.

During optimization, each Gaussian’s learnable attributes are updated via gradient descent under color supervision. An adaptive densification strategy further improves geometry and rendering quality by splitting Gaussians with high view-space gradients and removing those with low opacity or excessively large scale.

\subsection{Point Feature-Aware 3D Gaussian Splatting}

In this paper, we propose a point-wise feature-aware 3DGS method to improve novel view synthesis under few-shot settings. An overview of the proposed framework is illustrated in Fig.~\ref{fig:pipeline}. Given dense point clouds generated by VGGT~\cite{wang2025vggt} from a limited set of input images, we first extract multi-scale 2D features from the sparse views using Feature Pyramid Network(FPN). These features are then assigned to each 3D point by sampling from the corresponding image feature maps. To enhance appearance representation, we introduce a point interaction network that facilitates feature exchange among neighboring points. Finally, two lightweight MLPs decode the 3D Gaussian attributes—color $c$, opacity $\alpha$, and covariance $\Sigma$—from the enriched point-wise features. The details of each module are described below.

\noindent \textbf{(1) Dense Initialization.} 3D Gaussian Splatting (3DGS) is typically initialized using point clouds and camera poses derived from Structure-from-Motion (SfM) pipelines. However, traditional SfM-based methods often struggle to produce dense and reliable point clouds or accurate camera poses in challenging scenarios, particularly under sparse viewpoint coverage. To address this limitation, we introduce VGGT~\cite{wang2025vggt}, a stereo foundation model that performs end-to-end dense stereo matching from sparse multi-view inputs. VGGT can generate high-fidelity dense point clouds for each view while simultaneously estimating accurate camera poses. These initial reconstructions are further refined through bundle adjustment to improve both geometric and pose accuracy. Then these local point clouds are aligned and integrated into a globally consistent coordinate system using the estimated poses, ensuring geometric coherence across the reconstructed scene and providing a robust initialization for 3DGS.

\noindent \textbf{(2) Point Feature Representation.} 3D Gaussian Splatting  leverages spherical harmonics (SH) coefficients to efficiently model scene appearance, providing a compact representation of radiance fields. While SH is effective for capturing low-frequency lighting effects, it often fails to represent high-frequency details such as sharp shadows and specular highlights, limiting its applicability in scenes with complex illumination or fine structures.

To address this limitation, we introduce a novel appearance feature by fusing multi-scale image features from multiple input views to represent the color attributes of each Gaussian point. Specifically, we use a Feature Pyramid Network to extract multi-scale 2D feature maps $\{F_{i}^{1/4},F_{i}^{1/2}, F_{i}^{1} \}_{i=1}^N$ from $N$ sparse-view images $\{I_{i}\}_{i=1}^N$. Each 3D point is projected onto the image planes via the corresponding projection matrices, and the associated features are sampled from the multi-scale maps.

For the $j$-th point $X_j$, the sampled features from the $i$-th view are concatenated as $f_j^i \in \mathbb{R}^{56}$, comprising features of dimensions 8, 16, and 32 from the 1, 1/2, and 1/4 scale maps, respectively. The final appearance feature $f_j$ is obtained by aggregating multi-view features via element-wise variance:

\begin{equation}
f_j = \frac{1}{N} \sum_{i=1}^{N}(f_j^i - \bar{f}_j)^2
\end{equation}

where $\bar{f}_j$ is the average feature among input views. This fused feature $f_j$ is used to model the appearance of the Gaussian point with greater sensitivity to fine details.

\noindent \textbf{(3) Point-wise Feature Interaction}. Since 3D points in spatial proximity often share similar features, enabling interactions among neighboring points can enhance appearance representation and improve rendering quality. The self-attention mechanism is well-suited for point cloud processing, as it naturally accommodates their unordered, sparse, and irregular structure while capturing global context and supporting adaptive feature aggregation. Recent works in 3D scene understanding—such as semantic segmentation~\cite{robert2023efficient} and object classification~\cite{park2023self}—have demonstrated the effectiveness of transformer-based architectures with self-attention in modeling point-level relationships.

\begin{figure*}[htbp]
\begin{center}
    \includegraphics[width=0.6\textwidth]{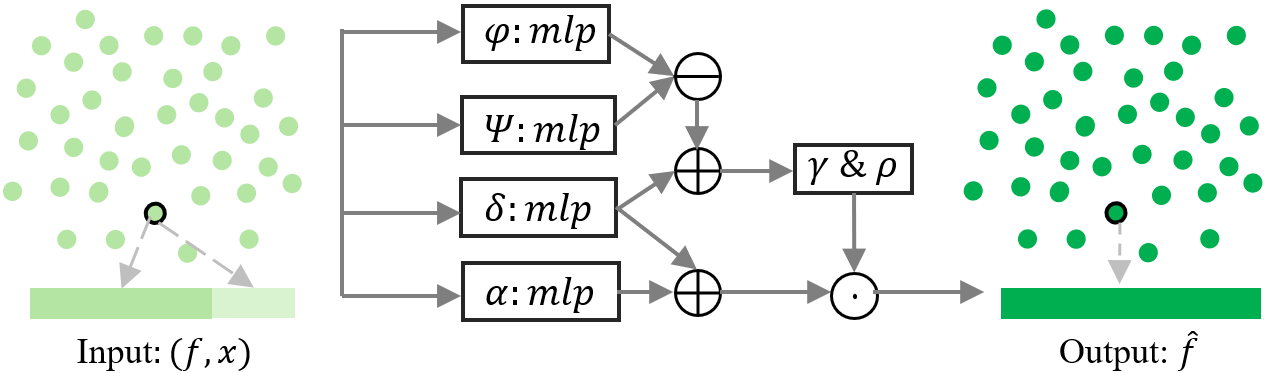}\\
    \caption{\textbf{Structure of point interaction.}The input is a set of point-wise feature vectors $f$ with associated 3D coordinates $x$. The point attention operators can  facilitates information exchange between these localized feature vectors and output the enhanced point features $\hat{f}$.} 
    \label{fig:pt}
\end{center}
\end{figure*}

Based on this argument, we appropriately design a Point Transformer network shown in Fig.\ref{fig:pt} that applies a local self-attention mechanism to process the acquired point clouds with appearance feature attributes. The input consists of point-wise features $f \in \mathbb{R}^{M \times 56}$, obtained from the previous step, and 3D point coordinates $x \in \mathbb{R}^{M \times 3}$.

As the unordered nature of point cloud data, new spatial position information is introduced to capture the spatial relationships between points.  In the point transformer, the input coordinates $x$ is fed into position encoding function $\delta$:

\begin{equation}
    \delta = \theta(x_i - x_j)
\end{equation}

where ${x_i}$ and ${x_j}$ is the 3D coordinates of point ${i}$ and ${j}$, and the encoding function $\theta$ is a multi-layer perceptron (MLP) consisting of two linear layers and a ReLU layer.

Then, the self-attention operator is defined as follow: 

\begin{equation}
    \hat{f_i}=\sum_{f_{j} \in \Omega(i)} \rho\left(\gamma\left(\varphi\left(f_i\right)-\psi\left(f_j\right)+\delta\right)\right) \odot\left(\alpha\left(f_j\right)+\delta\right)
\end{equation}

Here, $\Omega(i)\subseteq\Omega$ denotes the set of points in the $k$ nearest neighbors of $f_i$.  Specifically, for each point $f_i$, the KNN method is employed to identify its K-nearest neighbors, which collectively form the local neighborhood $\Omega(i)$.

Subsequently, the features of these neighboring points go through three point-wise feature transformations denoted by $\varphi$, $\psi$, and $\alpha$. Then, the result of subtracting $\varphi(f_i)$ from $\psi(f_j)$ and then adding $\delta$ undergoes a mapping function $\gamma$, an MLP with two linear layers and one ReLU.

Finally, the softmax function $\rho$ is further applied the output of $\gamma$ transformation and generate an attention weight to aggregate transformed features $\alpha(f_j)$ by the \textit{Hadamard product} $\odot$ and  outputs merged features $\hat{f_i}$.

Hence, this module can exploit local features via KNN, embeds spatial context through positional encoding, and combines attention-inspired operations with neural network layers to learn comprehensive appearance representations.

\noindent \textbf{(4) Gaussian Parameters.} After the point-wise features undergo multi-view fusion and feature interaction with spatial neighboring points, we can decode 3D Gaussian attributes $\{\mu_i, \Sigma_i, \alpha_i, c_i \}_{i=1}^M$ using two compact MLPs from the encoded features $\{\hat{f}_i\}_{i=1}^M$ for differentiable rendering. 

Each Gaussian center $\mu_i \in \mathbb{R}^3$ is obtained from the point cloud $\{X_{i}\}_{i=1}^M$. The first MLP network takes $X_i$ with 
positional encoding, current view direction $\phi \in \mathbb{R}^{3}$, and the encoded feature $\hat{f_i}$ as input, and predicts the color $c_i$ and the opacity $\alpha_i$ with $sigmoid$ function. The second MLP takes the same input as the first MLP and outputs covariance matrix $\Sigma_i$ by estimating its scaling $S_i$ with \textit{softplus} operator, rotation $R_i$ with \textit{normalization} operator.

\subsection{Optimization}  

Following original 3DGS, we firstly leverage a color reconstruction loss $\mathcal{L}_{color}$ consisting of $\mathcal{L}_1$ reconstruction loss and a structural-similarity loss $\mathcal{L}_{ssim}$:

\begin{equation}
    \mathcal{L}_{color} =   \lambda_1 \mathcal{L}_1 + \lambda_2 \mathcal L_{ssim}
\end{equation}

Recent methods~\cite{li2024dngaussian, chung2024depth} have shown that incorporating dense depth priors can effectively guide color-based optimization and regularize scene geometry across diverse datasets. Following this strategy, we introduce a depth regularization term $\mathcal L_{depth}$, defined as the $L_1$ distance between the rendered depth $D_{re}$ and the dense depth prior $\hat{D}$ predicted by a pre-trained monocular depth estimator~\cite{bhat2023zoedepth}.

\begin{equation}
    \mathcal{L}_{depth} =  ||D_{re} - \hat{D}||_1
\end{equation}

To promote smoothness in the rendered depth map, we introduce an additional depth smoothness term, defined as:
\begin{equation}
    \mathcal{L}_{smooth} =  e^{-\alpha_{1}\left|\nabla I_{gt}\right|}\left|\nabla D_{re}\right|+e^{-\alpha_{2}\left|\nabla^{2} I_{gt}\right|}\left|\nabla^{2} D_{re}\right|
\end{equation}
where $\alpha_1$ and $\alpha_2$ are set to 0.5, $I_{gt}$ is the input RGB image.

Finally, the total pixel-wise loss $\mathcal{L}_{total}$ is expressed as:
\begin{equation}
    \mathcal{L}_{total} = \mathcal{L}_{color}  + \lambda_3 \mathcal{L}_{depth} + \lambda_4 \mathcal{L}_{smooth}
\end{equation}
where $\lambda_1$, $\lambda_2$, $\lambda_3$ and $\lambda_4$ are set 0.8, 0.2, 0.05 and 0.067 by following the previous method~\cite{zhu2023fsgs}.

\section{Experiments}
\subsection{Datasets and Metrics}

\noindent \textbf{LLFF dataset.} LLFF~\cite{mildenhall2019local} contains eight scenes with forward-facing cameras. Following the same training and testing data split strategy as in~\cite{chung2024depth}, we selected every eighth image as the test set from each scene and
the remaining images were used for training. We randomly sampled 3 sparse views from the train set to train all methods and evaluate on the test set. Input resolution was downsampled by factors of 4× and 8×, resulting in dimensions of 504 × 378 and 1008 × 756, respectively.

\noindent \textbf{Mip-NeRF 360 dataset.} Mip-NeRF360~\cite{barron2022mip360} consists of 9 realistic scenes, including 5 outdoor and 4 indoor environments. Each scene features a complex central object or region, accompanied by a detailed background. Following the previous setup~\cite{zhu2023fsgs}, We selected 24 training views for comparison and test images were selected in the same manner as in the LLFF datasets. Input images were downsampled by 4× and 8× factors for fair comparison, respectively. 

\noindent \textbf{Deep Blending dataset.} We additionally evaluated the proposed method on the Deep Blending Dataset~\cite{hedman2018deep} with the bounded indoor scenes. Following 3DGS~\cite{kerbl20233d}, we selected \textit{Playroom} and \textit{DrJohnson} scenes from the dataset for performance comparison. We used 1/8 of all views for testing and randomly sampled 24 viewpoints from the remaining images in each scene for training.

\noindent \textbf{Baselines.} To comprehensively compare the proposed methods for novel view synthesis using sparse inputs, we compared our method with several representative NeRF-based methods~\cite{jain2021putting, niemeyer2022regnerf, wang2023sparsenerf, yang2023freenerf}, as well as the recent 3DGS-based approaches~\cite{zhu2023fsgs, xiong2023sparsegs, chung2024depth}. The best quantitative results are reported on these methods following the same experimental settings or referring to the results reported from the published papers.

\noindent \textbf{Evaluation Metrics.} To evaluate the quality in comparative view synthesis, we employed several commonly used metrics for evaluation: PSNR, LPIPS~\cite{zhang2018unreasonable} and SSIM~\cite{wang2004image}. Besides, FPS was also used for evaluating rendering speed.

\subsection{Implementation Details}

We implemented our method based on the official 3DGS codebase. For point-wise feature interaction, the number of neighboring points $K$ was set to 3. All models were trained on a single NVIDIA RTX 3090 GPU. Scene optimization was performed for 10,000 iterations across all scenes, with Gaussian densification starting at iteration 500 and repeated every 100 iterations. Two lightweight MLPs with a hidden dimension of 64 were used to decode Gaussian attributes from the learned point-wise features.

\subsection{Performance Evaluation}

\noindent \textbf{Results on LLFF dataset.} Quantitative results using 3 input views from the LLFF dataset are presented in Table \ref{tab:llff} with different resolutions. Our method outperformed NeRF-based methods significantly in all metrics. Compared to SparseNeRF~\cite{wang2023sparsenerf}, \textit{PSNR} was increased by 0.68dB and 0.9dB at two test resolutions. NeRF-style volumetric rendering requires large amount of computational resources, causing very slow inference. Within 3DGS-based methods, our method also surpassed the recent FSGS~\cite{zhu2023fsgs} by 0.32dB (reproduced) and 0.18dB (reported) in terms of \textit{PSNR}, though FPS declined slightly. This is because that the self-attention mechanisms used among Gaussian points increase computation costs to some degree.

\begin{figure*}[htbp]
\begin{center}
    \includegraphics[width=0.9\textwidth]{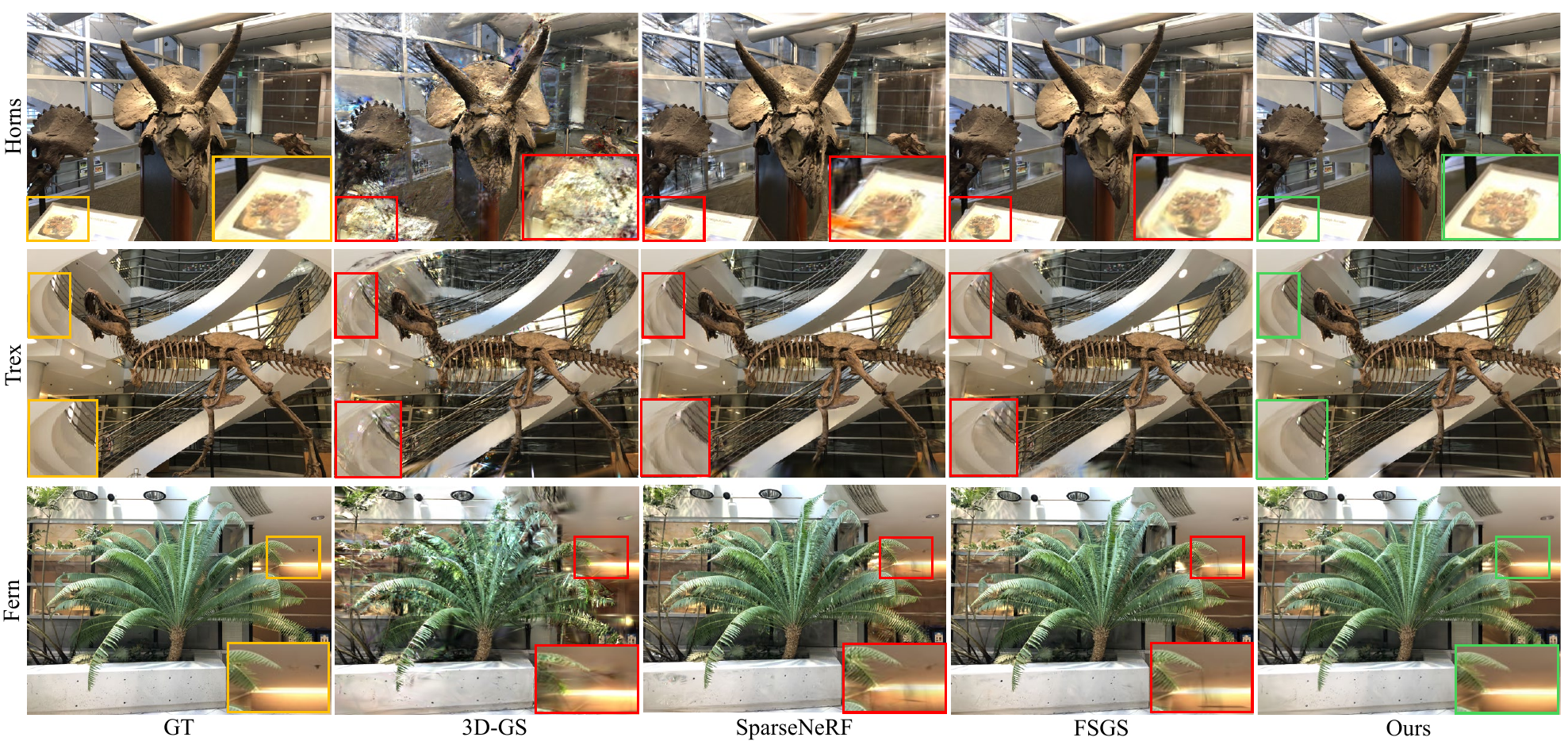}\\
    \caption{Visualized results on LLFF Datasets. Our method generates more realistic renderings with less blur in occluded and thin regions.
  }
\label{fig:llffpic}
\end{center}
\end{figure*}

\begin{table}[htbp]
\centering
\caption{Quantitative comparison on LLFF with 3 input views at 1/8 resolution.The best results are in \textbf{Bold} and the second best figures are in \underline{underlined.}}
\label{tab:llff}
\resizebox{1.0\columnwidth}{!}{

\begin{tabular}{llcccc}
\toprule
Category & Method & PSNR$\uparrow$ & SSIM$\uparrow$ & LPIPS$\downarrow$ & FPS$\uparrow$ \\
\midrule

\multirow{5}{*}{NeRF-based}
& Mip-NeRF~\cite{barron2021mip}         & 16.11 & 0.40 & 0.46  & 0.21 \\
& DietNeRF~\cite{jain2021putting}       & 14.94 & 0.37 & 0.49  & 0.14 \\
& RegNeRF~\cite{niemeyer2022regnerf}    & 19.08 & 0.59 & 0.336 & 0.21 \\
& FreeNeRF~\cite{yang2023freenerf}            & 19.63 & 0.61 & 0.31  & 0.21 \\
& SparseNeRF~\cite{wang2023sparsenerf}  & 19.86 & 0.62 & 0.33  & 0.21 \\
\midrule

\multirow{7}{*}{3DGS-based}
& Baseline (3DGS)                       & 16.89 & 0.46 & 0.41  & 354 \\
& DepthGS~\cite{chung2024depth}         & 17.17 & 0.50 & 0.34  & 374 \\
& DNGaussian~\cite{li2024dngaussian}    & 19.06 & 0.53 & 0.31  & 365 \\
& CoherentGS (repro)~\cite{paliwal2024coherentgs} & 20.13 & 0.64 & 0.32 & 403 \\
& \textcolor{gray}{FSGS(paper)}~\cite{zhu2023fsgs} & \textcolor{gray}{20.31} &\textcolor{gray}{0.65} &\textcolor{gray}{0.29} &\textcolor{gray}{458} \\
& FSGS (repro)                     & 20.22 & 0.67 & 0.27  & 448 \\
& \textbf{Ours}                         & \textbf{20.54} & \textbf{0.72} & \textbf{0.25} & \textbf{413} \\

\bottomrule
\end{tabular}
}
\end{table}

Rendering results are shown in Fig.\ref{fig:llffpic}, 3DGS has degraded quality due to insufficient view coverage. SparseNeRF and DepthGS utilize the depth priors to regularize scene geometry so to improve the rendering quality, yet still exhibited inadequate visual fidelity. FSGS adds additional Gaussians to the unobserved regions, thereby improving scene coverage and enhancing rendering quality. However, details in areas with highly complex geometry still failed to recover. For our method, each Gaussian point can learn separated appearance features from sparse view inputs and self-attention is further applied to facilitate the feature interaction with corresponding spatial neighboring points. The results verified that our method can capture more details and achieve better visual quality compared to other methods.

In addition, we compare our method with generalizable approaches such as MuRF~\cite{xu2024murf} and MVSplat~\cite{chen2024mvsplat} under the same input setting. Our method also outperforms both, achieving a higher PSNR on the LLFF dataset compared to MuRF (20.13 dB) and MVSplat (20.33 dB).

\noindent \textbf{Results on Mip-NeRF360 dataset.} Quantitative results of various methods on MipNeRF360 at two different resolutions are presented in Table \ref{tab:mipnerf}. As this dataset contains more complex outdoor scenarios, more dense view coverages are required. The proposed method demonstrated significant superiority compared with NeRF-based approaches. Compared to SparseNeRF, an improvement of 0.63dB in PSNR was achieved and FPS increased from 0.07 to 259  at 1/8 resolution. Within the recent few-shot 3DGS methods, our method also achieved competitive results at both resolutions when compared to the metrics reported in the FSGS paper. Our PointGS surpasses the reproduced results of FSGS under the same settings, by a 1.03dB increase in PSNR.
\begin{table}[htbp]
\centering
\caption{Quantitative comparison on Mip-NeRF360 with 24 input views. The best and the second best are \textbf{Bold} and \underline{underlined}.}
\label{tab:mipnerf}
\resizebox{1.0\columnwidth}{!}{
\begin{tabular}{llcccc}
\toprule
Category & Method & PSNR$\uparrow$ & SSIM$\uparrow$ & LPIPS$\downarrow$ & FPS$\uparrow$ \\
\midrule

\multirow{5}{*}{NeRF-based}
& Mip-NeRF~\cite{barron2021mip}         & 21.23 & 0.61 & 0.35 & 0.12 \\
& DietNeRF~\cite{jain2021putting}       & 20.21 & 0.56 & 0.39 & 0.05 \\
& RegNeRF~\cite{niemeyer2022regnerf}    & 22.19 & 0.64 & 0.34 & 0.07 \\
& FreeNeRF~\cite{yang2023freenerf}            & 22.78 & 0.69 & 0.32 & 0.07 \\
& SparseNeRF~\cite{wang2023sparsenerf}  & 22.85 & 0.69 & 0.32 & 0.07 \\
\midrule

\multirow{7}{*}{3DGS-based}
& Baseline (3DGS)                       & 20.42 & 0.58 & 0.34 & 208 \\
& DepthGS~\cite{chung2024depth}         & 21.20 & 0.61 & 0.29 & 211 \\
& DNGaussian~\cite{li2024dngaussian}    & 21.46 & 0.59 & 0.30 & 223 \\
& CoherentGS (repro)~\cite{paliwal2024coherentgs} & 22.33 & 0.61 & 0.35 & 227 \\
& \textcolor{gray}{FSGS(paper)} &\textcolor{gray}{23.70} &\textcolor{gray}{0.74} &\textcolor{gray}{0.22} &\textcolor{gray}{290} \\
& FSGS (repro)~\cite{zhu2023fsgs}   & 22.45 & 0.67 & 0.28 & 240 \\
& \textbf{Ours}                         & \textbf{23.48} & \textbf{0.72} & \textbf{0.24} & \textbf{231} \\

\bottomrule
\end{tabular}
}
\end{table}

\begin{figure*}[htbp]
\begin{center}
    \includegraphics[width=0.9\textwidth]{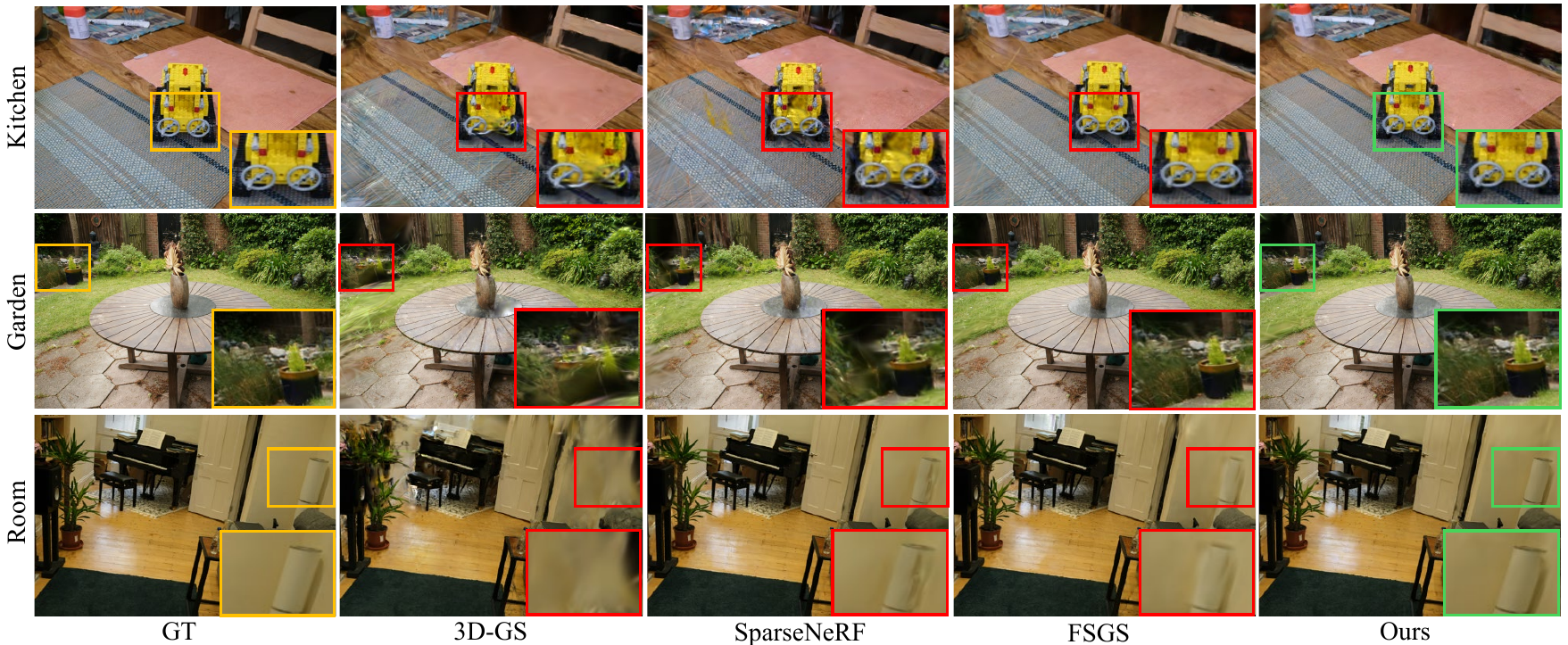}\\
    \caption{Visualized results on Mip-NeRF360 dataset.
  }
   \label{fig:mipnerfpic}
\end{center}
\end{figure*}
Visualized results on the MipNeRF360 are given in Fig.\ref{fig:mipnerfpic}.
Some scenes contain complex texture regions (such as abundantly growing herbaceous plants in \textit{Garden} scene) that are either invisible or observed from limited views, causing insufficient optimizations and blurry results. 3DGS and SparseNeRF failed to capture the intricate details of these scenes. FSGS achieved relatively good visual quality, but it still fell slightly short compared to our method. This is because our point-wise features sample from multiple images and hence can better model appearance features and capture more local details. Besides, point feature interaction using self-attention enables each 3D Gaussian point to learn more comprehensive geometric and appearance cues.

\begin{figure*}[htbp]
\begin{center}
    \includegraphics[width=0.9\textwidth]{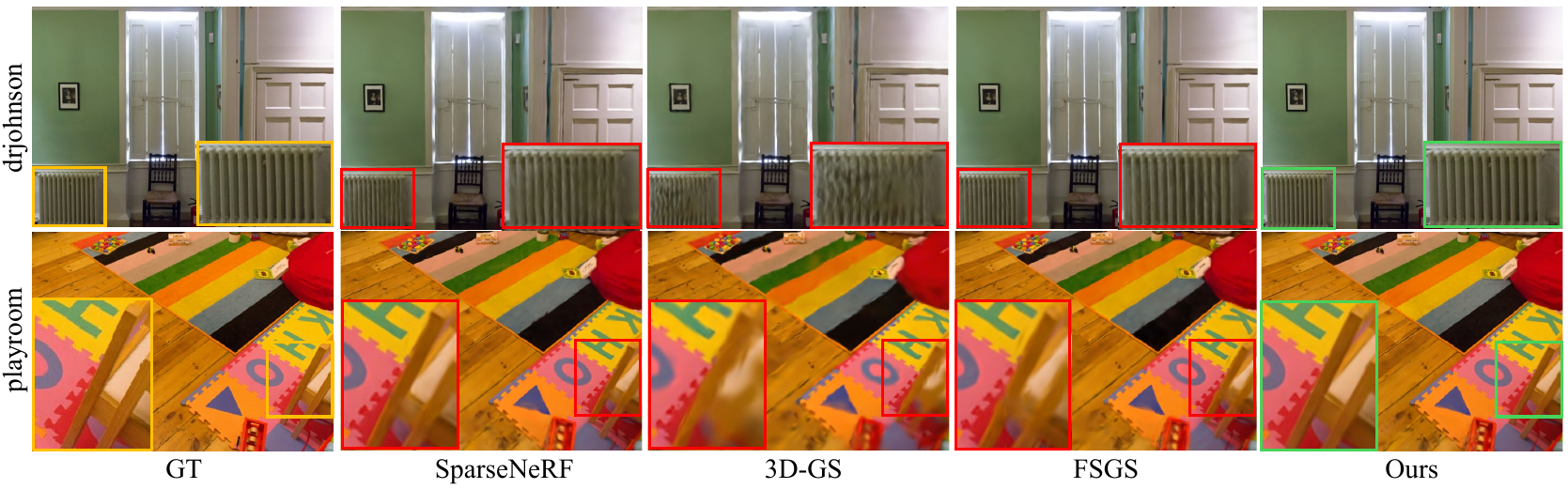}\\
    \caption{Qualitative comparison on Deep Blending with 24 input views.}
   \label{fig:db}
\end{center}
\end{figure*}

\begin{figure*}[ht]
\begin{center}
    \includegraphics[width=0.8\textwidth]{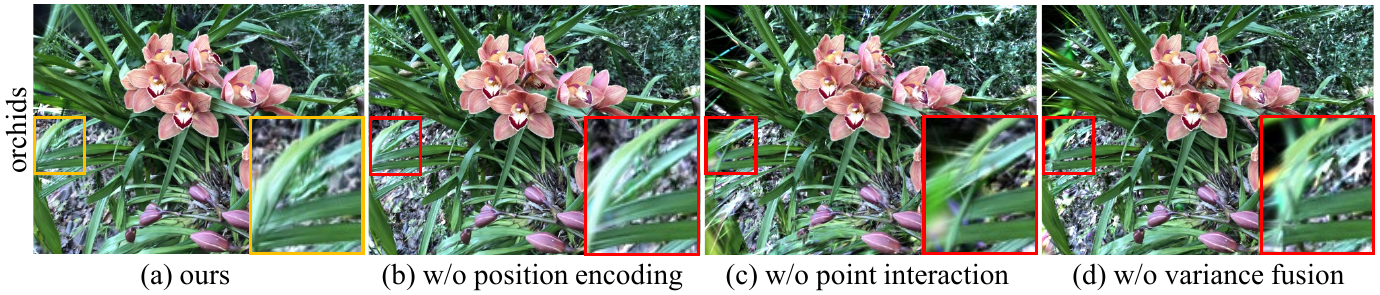}\\
    \caption{Ablation study on the components of PointGS.}
   \label{fig:abla}
\end{center}
\end{figure*}

\noindent \textbf{Results on Deep Blending dataset.} Quantitative comparison results are presented in Tab. \ref{tab:db} and visualized results are shown in Fig.\ref{fig:db}. The proposed PointGS consistently outperformed the previous state-of-the-art methods significantly in terms of \textit{PSNR} and exhibited high-quality visual results across various scenes.

\begin{table}[htbp]
\centering
\caption{Quantitative results on Deep Blending with 24 views. \textbf{Bold} and \underline{underlined} figures are the best and the second best.}
\resizebox{0.7\columnwidth}{!}{
\begin{tabular}{l|ccc}
\hline Method & \multicolumn{3}{c}{PSNR$\uparrow$ SSIM$\uparrow$  LPIPS$\downarrow$ } \\
\hline 
 RegNeRF~\cite{niemeyer2022regnerf} &19.65  &0.46   &0.38  \\
 SparseNeRF~\cite{wang2023sparsenerf} &20.79  &0.51  &0.34 \\
 \hline 
 Baseline (3DGS) &18.87  &0.48  &0.41  \\
 DNGaussian~\cite{li2024dngaussian} &22.16  &0.57  &0.32  \\
 CoherentGS~\cite{paliwal2024coherentgs} &22.85  &0.61  &0.31  \\
 FSGS~\cite{zhu2023fsgs} &\underline{23.46}  &\underline{0.67}  &\underline{0.29}  \\
 \hline 
 \textbf{Ours} & \textbf{24.32} & \textbf{0.79} & \textbf{0.25} \\
\hline
\end{tabular}
}
\label{tab:db}
\end{table}

\subsection{Ablation Studies}

To assess the influence of each component in the proposed method, we performed a series of ablation studies using the LLFF dataset with the 3-view input at 1/8 resolution. The quantitative and qualitative results of these studies are presented in the Tab. \ref{tab:abla}.

\begin{table}[htbp]
\centering
\caption{Ablation study on proposed components.}
\resizebox{0.8\columnwidth}{!}{
\begin{tabular}{l|cccc}
\hline Method & \multicolumn{3}{l}{PSNR  $\uparrow$  SSIM  $\uparrow$  LPIPS  $\downarrow$ } \\
\hline \textbf{Ours}  & \textbf{20.54} & \textbf{0.72} & \textbf{0.25} \\
\hline 
 w/o $\mathcal{L}_{smooth}$ & 20.49 & 0.71 & 0.26 \\
 w/o $\mathcal{L}_{depth}$ & 20.43 & 0.71 & 0.27 \\
 w/o position encoding &20.34  &0.68   &0.28  \\
 w/o point interaction &20.03  &0.62  &0.29  \\
 w/o variance fusion  & 19.95 & 0.61 & 0.29 \\

\hline
\end{tabular}
}
\label{tab:abla}
\end{table}

\noindent\textbf{Effect of depth information.} We sequentially removed depth smoothness loss and depth loss, resulting in lower performances. Without these two loss terms, the evaluation metrics continues to experience certain degrees of decline, decreasing from 20.54dB to 20.43dB in \textit{PSNR}.

\noindent\textbf{Effect of position encoding.} Removing positional encoding and using raw point coordinates directly leads to a drop in rendering quality, as shown in Fig.~\ref{fig:abla}(b), with the loss of high-frequency details in certain regions of the image.

\noindent\textbf{Effect of point interaction.} After removing the point interaction module leads to a noticeable performance drop, particularly in \textit{PSNR}, which decreases by 0.22 dB. As shown in Fig.~\ref{fig:abla}(c), the visual quality also degrades significantly. These results highlight the importance of this strategy.

\noindent\textbf{Effect of feature aggregation.} We replaced the variance-based fusion with simple averaging of multi-view image features. As shown in Tab.~\ref{tab:abla}, this leads to a slight drop in \textit{PSNR}. The performance degradation is attributed to the mean operation's inability to capture inter-view feature differences, unlike variance-based fusion.

\noindent\textbf{Color representation.} Compared with the learnable appearance features representation, using only SH coefficients for color prediction leads to a noticeable performance drop, with \textit{PSNR} decreasing from 20.54 to 20.36, \textit{SSIM} from 0.72 to 0.71, and \textit{LPIPS} increasing from 0.25 to 0.26.

\noindent\textbf{Number of Training Views.} We evaluate the impact of the number of training views on performance, as shown in Tab.~\ref{tab:trainview}, using the LLFF dataset with 2, 3, 5, 7, and 9 input views. As expected, increasing the number of views leads to notable improvements in performance metrics, particularly PSNR, due to broader scene coverage and better preservation of fine details. However, this comes at the cost of increased computational demands and a slight drop in inference speed.

\begin{table}[htbp]
\centering
\caption{Ablation study on the effects of input views.}
\resizebox{1.0\columnwidth}{!}{
\begin{tabular}{c|ccc|cc}
\hline
Methods & \multicolumn{1}{c}{PSNR$\uparrow$} & \multicolumn{1}{c}{SSIM$\uparrow$} & LPIPS$\downarrow$ & Memory(MB) & FPS \\ \hline
2-view  &18.89    &0.62      &0.36   &4189    &409     \\ 
3-view  &20.54    &0.72      &0.25   &4305    &395     \\ 
5-view  &22.61     & 0.78      &0.21      &4423        &386     \\ 
7-view  &23.57      &0.81       &0.19       &4541        &378     \\
9-view  &24.13      & 0.83     &0.18       &4598        &362     \\ \hline
\end{tabular}
}
\label{tab:trainview}
\end{table}

\noindent\textbf{Effects of K neighbors.} Tab.~\ref{tab:knn} shows
the impact of varying the number of neighborhood points ($K=0, 3, 6, 9$). 
\begin{table}[htbp]
\centering
\caption{Ablation study on the effects of $K$ neighbors.}
\resizebox{0.5\columnwidth}{!}{
\begin{tabular}{c|ccc}
\hline
K & \multicolumn{1}{c}{PSNR$\uparrow$} & \multicolumn{1}{c}{SSIM$\uparrow$} & LPIPS$\downarrow$  \\ \hline
0  &20.31   &0.66       &0.28         \\ 
3  &20.54   &0.72       &0.25          \\ 
6  &20.62   &0.77       &0.22          \\ 
9  &20.71   &0.79       &0.22         \\ 
\hline
\end{tabular}
}
\label{tab:knn}
\end{table}
Rendering quality improves with larger $K$, as each point can exchange appearance information with more neighbors and learn richer features.

\section{Conclusions and Limitations}

In this paper, we proposed a few-shot novel view synthesis method based on 3D Gaussian Splatting. To better model scene appearance, we introduced point-wise appearance features aggregated from multi-view images, and further enhanced their representation through a self-attention-based point interaction module. Extensive experiments demonstrate that our approach significantly outperforms NeRF-based methods and achieves competitive performance compared to state-of-the-art 3DGS approaches under the same few-shot setting. However, the proposed point-wise feature interaction introduces additional computational overhead and may slightly reduce inference speed. Moreover, our method struggles to generalize to unseen viewpoints in the presence of occlusions. In future work, we plan to explore efficient sparse attention mechanisms to reduce computational costs, and incorporate richer pixel-level and spatial cues to further improve novel view synthesis quality.

\newpage
{
    \small
    \bibliographystyle{ieeenat_fullname}
    \bibliography{main}
}

\end{document}